\documentclass{article}
\usepackage{spconf,amsmath,graphicx,hyperref}
\usepackage{multirow}

\title{SAM-LLM: INTERPRETABLE LANE CHANGE TRAJECTORY PREDICTION VIA PARAMETRIC FINETUNING}

\name{Zhuo Cao\sthanks{Thanks to XYZ agency for funding.}\textsuperscript{\dag}
      \quad Yunxiao Shi\textsuperscript{\ddag}
      \quad Min Xu\textsuperscript{\ddag}}
\address{\textsuperscript{\dag}\,University of Queensland, School of Civil Engineering\\
         \textsuperscript{\ddag}\,University of Technology Sydney, Faculty of Engineering and Information Technology}

\begin{document}
%
\maketitle
\begin{abstract}
This work introduces SAM-LLM, a novel hybrid architecture that bridges the gap between the contextual reasoning of Large Language Models (LLMs) and the physical precision of kinematic lane change models for autonomous driving. The system is designed for interpretable lane change trajectory prediction by finetuning an LLM to output the core physical parameters of a trajectory model instead of raw coordinates. For lane-keeping scenarios, the model predicts discrete coordinates, but for lane change maneuvers, it generates the parameters for an enhanced Sinusoidal Acceleration Model (SAM), including lateral displacement, maneuver duration, initial lateral velocity, and longitudinal velocity change. This parametric approach yields a complete, continuous, and physically plausible trajectory model that is inherently interpretable and computationally efficient, achieving an 80\% reduction in output size compared to coordinate-based methods. The SAM-LLM achieves a state-of-the-art overall intention prediction accuracy of 98.73\%, demonstrating performance equivalent to traditional LLM predictors while offering significant advantages in explainability and resource efficiency.
\end{abstract}
\begin{keywords}
Large Language Models, Finetuning, Lane Change Prediction, Physical Explainability, Resource Efficiency
\end{keywords}
\section{Introduction}
\label{sec:intro}



Lane change prediction is fundamental to autonomous driving, requiring accurate anticipation of driver intent and vehicle motion for safe navigation in dynamic traffic environments \cite{song2021surrounding, bi2025lane}. Over the past decade, mainstream approaches have relied on discriminative deep learning architectures—LSTMs \cite{xin2018intention, scheel2022recurrent}, CNNs \cite{mozaffari2022early}, and Transformers \cite{gao2023dual}—to model vehicle interactions and predict trajectories through coordinate regression. With the rapid advancement of large language models (LLMs) and their demonstrated capabilities in zero-shot decision making, contextual reasoning, and cross-domain knowledge generalization, a new promising research frontier in lane change prediction is emerging: LLM-based  lane change prediction method represents a paradigm shift toward generative methodologies. Early explorations suggest that this possibility is more than speculative. 







The LC-LLM approach \cite{peng2025lc} pioneered the application of LLMs to lane change prediction, using Chain-of-Thought (CoT) reasoning \cite{wei2022chain} to generate natural language explanations alongside intention and trajectory predictions. While achieving significant improvements in interpretability, coordinate sequence-based LLM approaches face inherent challenges: they require the model to generate precise numerical sequences for trajectory points, leading to computational inefficiency and potential accumulation of small errors across multi-point predictions.
A fundamental limitation of existing approaches lies in the representation of lane change trajectories. Current methods output discrete coordinate sequences that lack physical grounding and require extensive data to capture the underlying kinematic principles governing vehicle motion \cite{lan2024traj,peng2025lc}. This representation is neither computationally efficient nor inherently interpretable, as the physical meaning of individual trajectory points remains opaque.


In this paper, we propose SAM-LLM, a novel hybrid architecture that addresses these limitations by leveraging LLMs to generate physically meaningful parameters rather than raw coordinates. Our approach employs an enhanced Sinusoidal Acceleration Model (SAM) \cite{jula2002collision, wang2015driving} for lane change maneuvers, where the LLM outputs key physical parameters: lateral displacement, maneuver duration, initial lateral velocity, and longitudinal velocity change. For lane-keeping scenarios, the system maintains coordinate-based predictions, creating a hybrid strategy that optimally balances efficiency and accuracy.
This parametric approach offers several key advantages: (1) Physical interpretability - each parameter has clear kinematic meaning, enabling direct analysis of driving behavior; (2) Computational efficiency - achieving 80\% reduction in output size compared to coordinate-based methods; (3) Trajectory completeness - generating continuous, smooth trajectories that extend beyond discrete prediction horizons; (4) Model consistency - ensuring physically plausible predictions through parametric constraints.
Our contributions include: (1) Introduction of the first parametric LLM approach for lane change prediction, bridging contextual reasoning with kinematic modeling; (2) Development of a hybrid fine-tuning strategy combining coordinate and parameter prediction; (3) Achievement of state-of-the-art performance on the highD \cite{krajewski2018highd} dataset with 98.73\% intention prediction accuracy; (4) Demonstration of significant computational efficiency gains while maintaining prediction quality.

\section{Problem Formulation and Methodology}
\label{sec:method}

\begin{figure}[htb]
\centering
\includegraphics[width=\columnwidth]{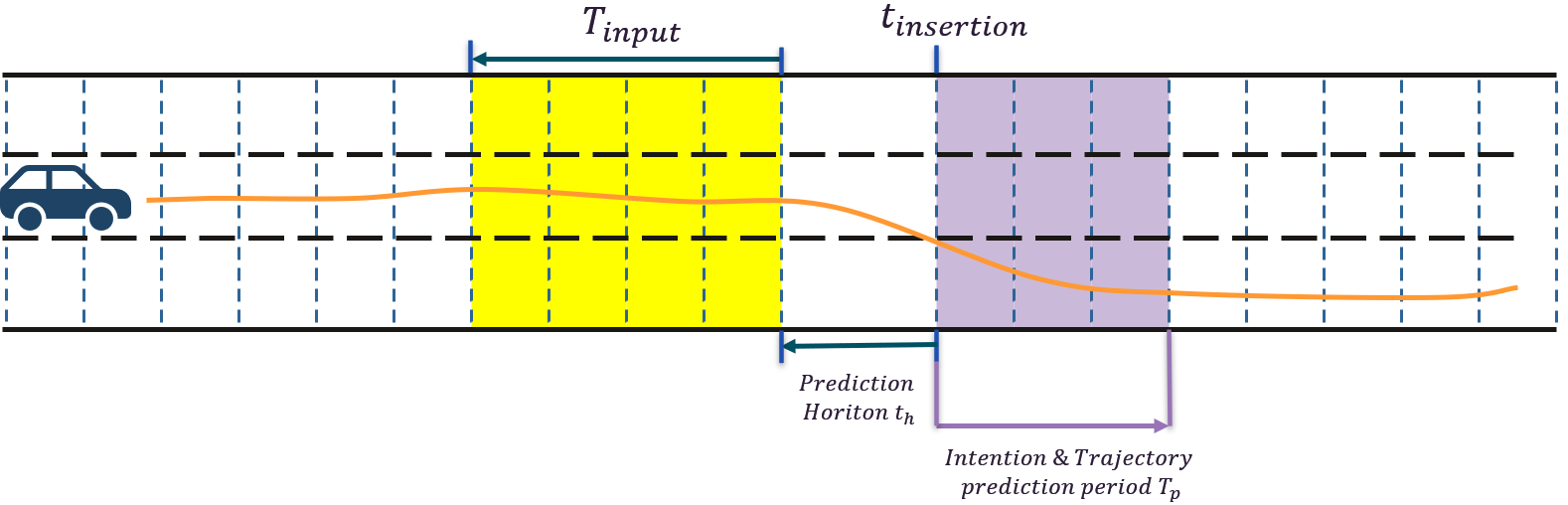}
\caption{Problem formulation: Lane change prediction using historical observations ($T_{input}$) to predict intention and trajectory during prediction period ($T_p$). The insertion point ($t_{insertion}$) marks the transition from observation to prediction phases. The prediction horizon $t_h$ represents the time gap between the available observation time and the start of prediction period $T_p$.}
\label{fig:problem_formulation}
\end{figure}

Figure~\ref{fig:problem_formulation} illustrates the temporal framework of lane change intention and trajectory predictions. Given a sequence of historical observations during the input window $T_{input}$, our objective is to predict both the lane change intention and future trajectory of the target vehicle during the prediction period $T_p$.

Formally, let $\mathbf{H} = \{\mathbf{s}_{t-T_{input}}, ..., \mathbf{s}_{t-1}\}$ represent the historical vehicle states during the input window, where each state $\mathbf{s}_i$ contains position, velocity, acceleration, and contextual information. At the insertion point $t_{insertion}$, our model generates the lane change intention $I \in \{0, 1, 2\}$ representing keep lane, left lane change, and right lane change respectively, along with future trajectory representation $\mathbf{T}$ describing vehicle motion during prediction period $T_p$.

\subsection{SAM-LLM: Hybrid Parametric-Coordinate Prediction}

While recent LC-LLM approaches \cite{peng2025lc} demonstrate the effectiveness of LLMs for lane change prediction through natural language reasoning, they rely on coordinate-based trajectory representation which lacks physical interpretability and computational efficiency. Our key innovation lies in a hybrid trajectory representation strategy:

\begin{equation}
\mathbf{T} = \begin{cases}    
\mathbf{T}_{coord} = \{(x_1, y_1), ..., (x_4, y_4)\} & \text{if } I = 0 \\
\mathbf{T}_{param} = \{W, D, v_0, \Delta v_x\} & \text{if } I \in \{1, 2\}
\end{cases}
\end{equation}

where $\mathbf{T}_{coord}$ represents discrete trajectory coordinates for lane keeping scenarios, while $\mathbf{T}_{param}$ contains physically meaningful SAM parameters for lane change maneuvers: lateral displacement ($W$), maneuver duration ($D$), initial lateral velocity ($v_0$), and longitudinal velocity change ($\Delta v_x$).

\begin{figure}[htb]
\centering
\includegraphics[width=\columnwidth]{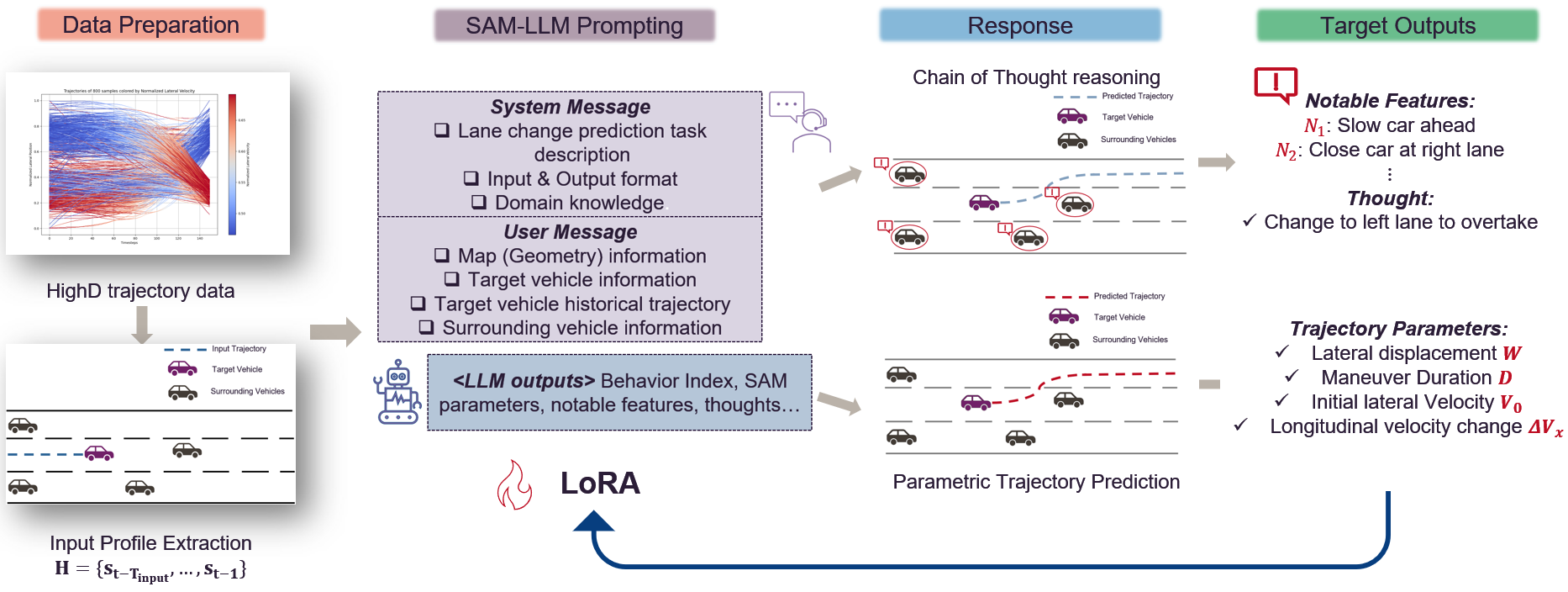}
\caption{SAM-LLM architecture: SAM-LLM framework consists of three main components: (1) a multi-modal input encoder that converts heterogeneous driving data into natural language prompts, (2) a fine-tuned Llama-2-7B backbone for spatial reasoning, and (3) a hybrid output decoder generating either coordinate sequences or SAM parameters based on predicted intentions.}
\label{fig:sam_llm_architecture}
\end{figure}

\subsection{Chain-of-Thought Prompting for Interpretability}

To enhance interpretability, we adopt the Chain-of-Thought (CoT) framework \cite{wei2022chain} for lane change prediction tasks \cite{peng2025lc}, where the model is fine-tuned to explicitly verbalize its reasoning process before providing a final prediction. This process involves generating a Thought section that first identifies salient notable features from the driving context (e.g., a blocked lane ahead, significant lateral movement) and then synthesizes them into a potential behavior (e.g., overtaking). This intermediate reasoning step serves as a direct justification for the Final Answer, which contains the intention I and trajectory representation T. The CoT approach forces the model into a structured reasoning pathway, improving robustness while yielding a human-readable explanation for each prediction.

\subsection{Enhanced Sinusoidal Acceleration Model (SAM)}

The Sinusoidal Acceleration Model (SAM) from transport studies \cite{jula2002collision, mullakkal2020empirics, wang2015driving} offers significant advantages over coordinate-based approaches through its physically interpretable parameters and smooth trajectory generation based on human driving behavior. However, the original SAM cannot be directly applied to our prediction scenario. The classical SAM models complete lane change trajectories from maneuver initiation to completion:

\begin{equation}
y(t) = y_0 + \frac{-W}{2\pi}\sin\left(\frac{2\pi(t-t_{start})}{D}\right) + \frac{W(t-t_{start})}{D}
\end{equation}

Since our SAM-LLM makes predictions at the lane boundary crossing point ($t_{insertion}$) rather than at lane change initiation, we require only the second half of the trajectory. We therefore propose a modified SAM that approximates the post-boundary trajectory segment, incorporating the lateral velocity at boundary crossing ($v_0$) as an initial condition:
\begin{equation}
y(t) = v_0 t + \frac{W - 2v_0 D}{\pi} \sin\left(\frac{\pi t}{2D}\right)
\end{equation}

The longitudinal motion incorporates velocity changes as linear transitions:
\begin{equation}
v_x(t) = v_{x,0} + \frac{\Delta v_x \cdot t}{D}
\end{equation}

This modified SAM formulation addresses the specific timing of our prediction task by ensuring boundary conditions: zero lateral acceleration at maneuver start and end ($a_y(0) = a_y(D) = 0$), producing physically plausible trajectories consistent with natural driving behavior. In addition, Each parameter has clear physical significance: lateral displacement $W$ (typically 3.5-4.0m), maneuver duration $D$ (3-6 seconds), initial lateral velocity $v_0$ (extracted at boundary crossing), and longitudinal velocity change $\Delta v_x$ capturing acceleration behavior.

\subsection{Hybrid Fine-tuning Strategy}

We implement a hybrid fine-tuning approach that handles both coordinate and parametric outputs within a unified language modeling framework. The training data preparation involves: (1) coordinate extraction for lane-keeping scenarios; (2) SAM parameter fitting for lane-change maneuvers using least-squares optimization on ground truth trajectories; and (3) CoT reasoning generation through rule-based feature extraction.

For lane-changing scenarios ($I \in \{1,2\}$), SAM parameters are fitted via:
\begin{equation}
\hat{\boldsymbol{\theta}} = \arg\min_{\boldsymbol{\theta}} \sum_{t \in T_p} ||y_{gt}(t) - y_{SAM}(t; \boldsymbol{\theta})||^2
\end{equation}
where $\boldsymbol{\theta} = \{W, D, v_0, \Delta v_x\}$ represents the fitted parameters.

The complete model output is structured as a single text sequence $\mathbf{S} = [\mathbf{P}, \mathbf{C}, I, \mathbf{T}]$, where $\mathbf{P}$ represents input prompt tokens, $\mathbf{C}$ the CoT reasoning, $I$ the intention, and $\mathbf{T}$ the trajectory representation (coordinates or parameters). We apply standard causal language modeling:
\begin{equation}
\mathcal{L} = -\sum_{i=1}^{N} \sum_{j=|\mathbf{P}|+1}^{|\mathbf{S}|} \log P(s_j^{(i)} | s_{<j}^{(i)}, \mathbf{P}^{(i)}; \boldsymbol{\phi})
\end{equation}

This unified objective implicitly learns intention classification, CoT generation, and trajectory representation simultaneously. We employ LoRA fine-tuning \cite{hu2022lora} with rank $r=64$, targeting attention projection layers of Llama-2-7B, achieving parameter-efficient adaptation while maintaining prediction quality across both representation formats.

\section{Experimental Results}
\label{sec:results}

\begin{table*}[t]   
\centering
\caption{Performance and efficiency comparison on highD dataset.}
\label{tab:results}
\small   
\begin{tabular}{l ccc ccc ccc c}   
\hline
\multirow{2}{*}{Method} & \multicolumn{3}{c}{Keep Lane (1836)} & \multicolumn{3}{c}{Left LC (428)} & \multicolumn{3}{c}{Right LC (502)} & \multicolumn{1}{c}{\begin{tabular}[c]{@{}c@{}}Process\\ Time (ms)\end{tabular}} \\
\cline{2-4} \cline{5-7} \cline{8-10}
& Acc & Lat & Lon & Acc & Lat & Lon & Acc & Lat & Lon & \\
\hline
LC-LLM (4-pt) & 98.97 & 0.167 & 1.047 & 97.20 & 0.301 & 1.483 & 98.41 & 0.289 & 1.130 & 915.8 \\
LC-LLM (20-pt) & 99.13 & 0.186 & 0.977 & 97.43 & 0.330 & 1.375 & 98.61 & 0.310 & 1.060 & 1627.8 \\
\textbf{SAM-LLM} & \textbf{99.07} & \textbf{0.165} & 1.045 & \textbf{97.43} & \textbf{0.286} & 1.539 & \textbf{98.61} & \textbf{0.264} & 1.223 & \textbf{747.3} \\
\hline
\end{tabular}
\end{table*}

\begin{table*}[t]   
\centering
\caption{Point-by-point error analysis across all intention classes (RMSE values in meters).}
\label{tab:pointwise_errors}
\scriptsize   
\begin{tabular}{cc|ccc|ccc|ccc}   
\hline
\multirow{2}{*}{\begin{tabular}[c]{@{}c@{}}Time\\Point\end{tabular}} & \multirow{2}{*}{Metric} & \multicolumn{3}{c|}{Keep Lane (1836)} & \multicolumn{3}{c|}{Left Lane Change (428)} & \multicolumn{3}{c}{Right Lane Change (502)} \\
\cline{3-11}
& & LC-LLM & LC-LLM & \textbf{SAM-LLM} & LC-LLM & LC-LLM & \textbf{SAM-LLM} & LC-LLM & LC-LLM & \textbf{SAM-LLM} \\
& & (4-pt) & (20-pt) & & (4-pt) & (20-pt) & & (4-pt) & (20-pt) & \\
\hline
\multirow{2}{*}{1s} & Lateral & 0.066 & 0.065 & \textbf{0.062} & 0.122 & 0.126 & \textbf{0.123} & 0.111 & 0.118 & \textbf{0.113} \\
& Longitudinal & 0.238 & 0.209 & \textbf{0.235} & 0.332 & 0.293 & 0.363 & 0.238 & 0.213 & 0.261 \\
\hline
\multirow{2}{*}{2s} & Lateral & 0.130 & 0.133 & \textbf{0.122} & 0.244 & 0.255 & \textbf{0.242} & 0.223 & 0.230 & \textbf{0.217} \\
& Longitudinal & 0.595 & 0.550 & 0.587 & 0.839 & 0.777 & 0.883 & 0.619 & 0.586 & 0.655 \\
\hline
\multirow{2}{*}{3s} & Lateral & 0.181 & 0.198 & \textbf{0.171} & 0.330 & 0.354 & \textbf{0.324} & 0.312 & 0.326 & \textbf{0.303} \\
& Longitudinal & 1.072 & 1.027 & 1.062 & 1.522 & 1.452 & 1.570 & 1.150 & 1.115 & 1.174 \\
\hline
\multirow{2}{*}{4s} & Lateral & 0.219 & 0.256 & \textbf{0.208} & 0.390 & 0.419 & \textbf{0.366} & 0.378 & 0.411 & \textbf{0.350} \\
& Longitudinal & 1.649 & 1.611 & 1.632 & 2.343 & 2.265 & 2.630 & 1.798 & 1.763 & 2.463 \\
\hline
\end{tabular}
\end{table*}

\subsection{Experimental Setup}
We evaluate SAM-LLM on the highD dataset using 2,766 test samples from highway scenarios, comparing against LC-LLM baselines with both 4-point and 20-point trajectory outputs. All models use an identical Llama-2-7B backbone and LoRA fine-tuning for a fair comparison.

\subsection{Results and Analysis}
As shown in Table~\ref{tab:results}, SAM-LLM achieves a state-of-the-art overall intention prediction accuracy of 98.73\% while demonstrating competitive trajectory prediction performance. The hybrid approach maintains consistently high accuracy across all intention classes: 99.07\% for lane keeping, 97.43\% for left lane changes, and 98.61\% for right lane changes.

\textbf{Point-by-Point Error Analysis:} Table~\ref{tab:pointwise_errors} provides detailed temporal error analysis across the 4-second prediction horizon. SAM-LLM demonstrates superior lateral trajectory prediction performance, consistently achieving the lowest RMSE values across all time points and intention classes. For lateral errors, our method shows improvements ranging from 6.1\% at 1-second predictions to 19.9\% at 4-second predictions compared to LC-LLM baselines. This superior long-term accuracy demonstrates the effectiveness of the physically-grounded parametric approach in maintaining trajectory precision over extended prediction horizons.

Notably, the error progression analysis reveals that while all methods experience increasing errors with longer prediction horizons (as expected), SAM-LLM maintains more stable lateral prediction accuracy. For lane change scenarios, the lateral RMSE increases from 0.123m at 1-second to 0.366m at 4-second for left lane changes, representing a more gradual degradation compared to coordinate-based approaches.

\textbf{Computational Efficiency:} SAM-LLM demonstrates remarkable computational advantages. By generating 4 physical parameters instead of 20 coordinates, it reduces output dimensionality by 80\%, leading to a 54\% inference speedup (747.3ms vs 1627.8ms for the 20-point baseline). This efficiency gain is particularly valuable for real-time autonomous driving applications where computational resources are constrained.

\begin{figure}[th]
\centering
\includegraphics[width=\columnwidth]{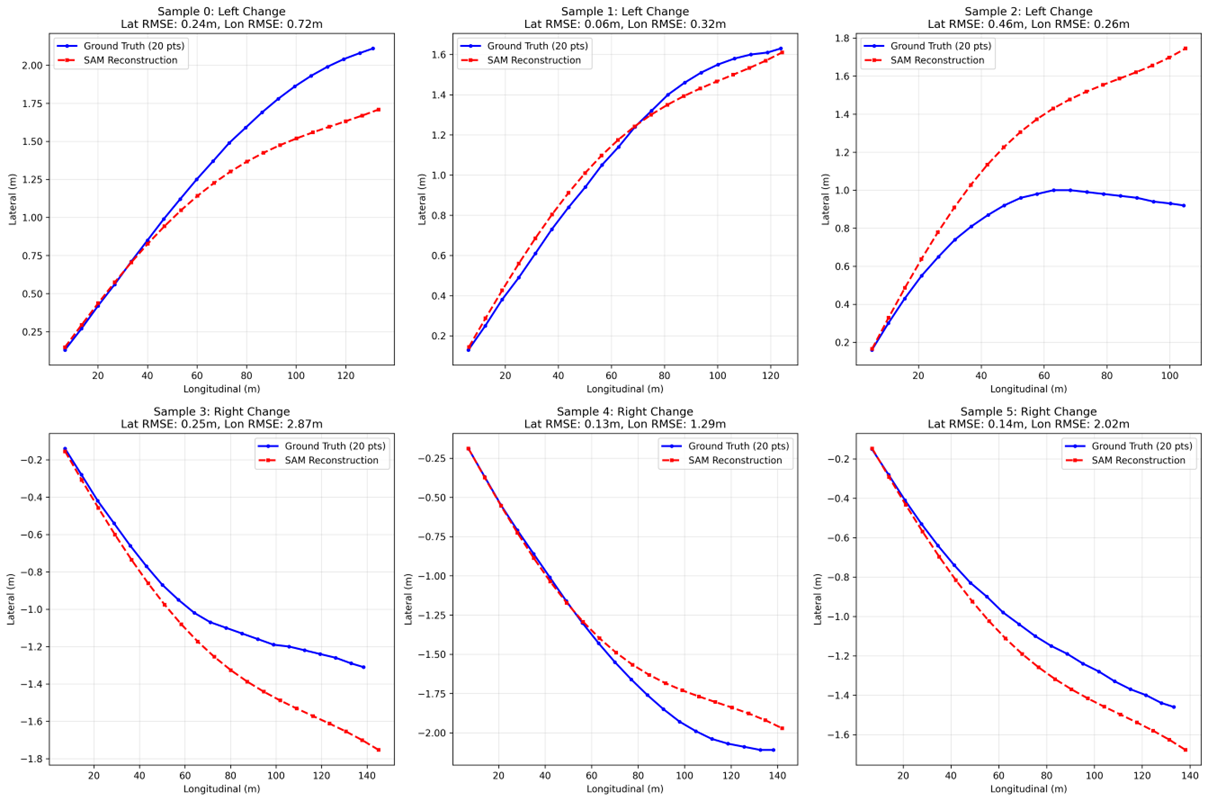}
\caption{Qualitative comparison of SAM-reconstructed trajectories (red, dashed) against ground truth (blue, solid) for various lane change samples. Our parametric approach generates smooth and physically plausible paths that closely follow the real-world data.}
\label{fig:traj_samples}
\end{figure}

\begin{figure}[th]
\centering
\includegraphics[width=\columnwidth]{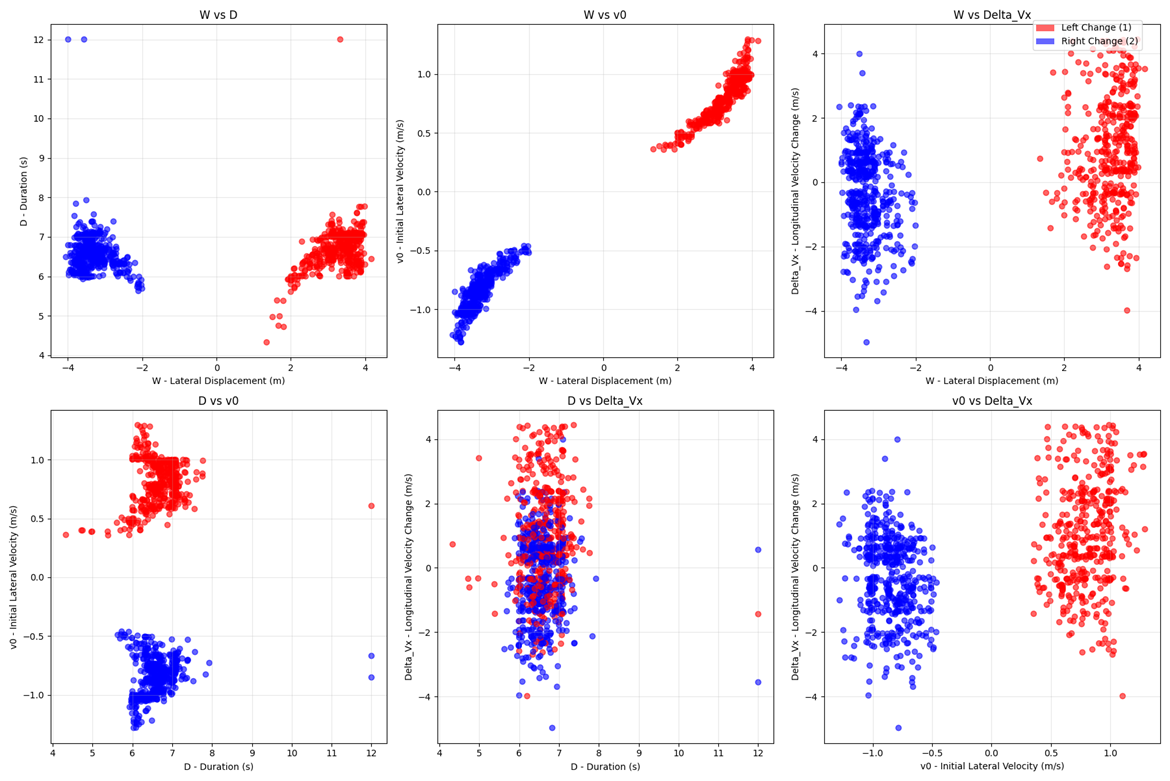}
\caption{Distributions of predicted SAM parameters for left (red) and right (blue) lane changes. The clear separation and tight clustering in plots like W vs D and W vs v0 demonstrate that the model learns physically meaningful and consistent driving behaviors.}
\label{fig:param_dist}
\end{figure}

\textbf{Physical Interpretability:} Figure~\ref{fig:param_dist} illustrates the interpretability advantages of our parametric approach. The predicted SAM parameters form distinct, physically meaningful clusters for left and right lane changes. The clear separation in parameter distributions (e.g., lateral displacement $W$ vs duration $D$) confirms that the model learns consistent and realistic driving behaviors, providing transparent insights that coordinate-based methods cannot offer. This interpretability enables direct analysis of driving patterns, such as typical lane change durations (3-6 seconds) and lateral displacements (3.5-4.0m), aligning with established transportation engineering knowledge. Figure~\ref{fig:traj_samples} provides qualitative validation, showing that trajectories reconstructed from predicted SAM parameters closely align with ground truth data while producing smooth, physically plausible paths that extend continuously beyond the discrete prediction horizon.

\section{Conclusion}
\label{sec:conclusion}

This paper introduces SAM-LLM, a novel hybrid architecture that bridges Large Language Model reasoning with physically-grounded trajectory modeling for lane change prediction. By fine-tuning LLMs to output SAM parameters rather than raw coordinates, our approach achieves state-of-the-art performance (98.73\% intention accuracy) while providing 80\% reduction in output dimensionality, 54\% inference speedup, and inherent physical interpretability through meaningful parameter clustering.

The parametric approach enables direct analysis of driving behaviors through physically interpretable parameters, offering crucial insights for safety-critical autonomous driving systems. Future work will focus on enhanced longitudinal dynamics modeling and extension to complex urban driving scenarios. This framework establishes a foundation for physics-informed trajectory prediction with language models, demonstrating the potential for integrating domain knowledge with modern LLMs for more interpretable and efficient autonomous driving systems.

\vfill\pagebreak
\bibliographystyle{IEEEbib}
\bibliography{strings,refs}

\end{document}